\documentclass[letterpaper]{article} 
\usepackage{aaai24}  
\usepackage{times}  
\usepackage{helvet}  
\usepackage{courier}  
\usepackage[hyphens]{url}  
\usepackage{graphicx} 
\usepackage{natbib}  
\usepackage{caption} 
\usepackage{algorithm}
\usepackage{algorithm}

\usepackage{graphicx}
\usepackage{amsmath, amssymb} 
\usepackage{bm}
\usepackage{textcomp}
\usepackage{gensymb}
\usepackage{pifont}
\usepackage[T1]{fontenc}
\usepackage{multirow}
\usepackage{makecell}
\usepackage{xspace}
\usepackage{enumitem}
\usepackage{url}
\usepackage{enumerate}
\usepackage{algorithm}
\usepackage[noend]{algpseudocode} 
\makeatletter
\DeclareMathOperator*{\argmax}{arg\,max}
\DeclareMathOperator*{\argmin}{arg\,min}
\DeclareRobustCommand\onedot{\futurelet\@let@token\@onedot}
\def\@onedot{\ifx\@let@token.\else.\null\fi\xspace}
\def\eg{\emph{e.g}\onedot} 
\def\ie{\emph{i.e}\onedot}

\def\etal{\emph{et al}\onedot}
\def\ve{\boldsymbol}

\newcommand{\cmark}{\ding{51}}%
\newcommand{\xmark}{\ding{55}}%
\newcommand\ChangeRT[1]{\noalign{\hrule height #1}}
\makeatother

\usepackage{newfloat}
\usepackage{listings}
\DeclareCaptionStyle{ruled}{labelfont=normalfont,labelsep=colon,strut=off} 
\floatstyle{ruled}
\newfloat{listing}{tb}{lst}{}
\floatname{listing}{Listing}
\nocopyright 

\title{Training with Product Digital Twins for AutoRetail Checkout}
\author{
    Yue Yao\textsuperscript{\rm 1}\equalcontrib\thanks{This work is partially done when Yue has an
internship at NVIDIA.},
    Xinyu Tian\textsuperscript{\rm 1}\equalcontrib,
    Zheng Tang$^{2}$, Sujit Biswas$^{2}$, Huan Lei$^{3}$, Tom Gedeon\textsuperscript{\rm 4}, Liang Zheng\textsuperscript{\rm 1} 
}
\affiliations{
    \textsuperscript{\rm 1}The Australian National University\\
    \textsuperscript{\rm 2}NVIDIA\\
    \textsuperscript{\rm 3}The University of Adelaide\\
    \textsuperscript{\rm 4}Curtin University\\

}

\usepackage{bibentry}

\begin{document}

\maketitle

\begin{abstract}
Automating the checkout process is important in smart retail, where users effortlessly pass products by hand through a camera, triggering automatic product detection, tracking, and counting. In this emerging area, due to the lack of annotated training data, we introduce a dataset comprised of product 3D models, which allows for fast, flexible, and large-scale training data generation through graphic engine rendering. Within this context, we discern an intriguing facet, because of the user ``hands-on'' approach, bias in user behavior leads to distinct patterns in the real checkout process. The existence of such patterns would compromise training effectiveness if training data fail to reflect the same. To address this user bias problem, we propose a training data optimization framework, \ie, training with digital twins (DtTrain). Specifically, we leverage the product 3D models and optimize their rendering viewpoint and illumination to generate  ``digital twins'' that visually resemble representative user images. These digital twins, inherit product labels and, when augmented, form the Digital Twin training set (DT set). Because the digital twins individually mimic user bias, the resulting DT training set better reflects the characteristics of the target scenario and allows us to train more effective product detection and tracking models\footnote{Counting is an inherent outcome once all products are successfully detected and tracked.}. In our experiment, we show that DT set outperforms training sets created by existing dataset synthesis methods in terms of counting accuracy. Moreover, by combining DT set with pseudo-labeled real checkout data, further improvement is observed. The code is available at \url{https://github.com/yorkeyao/Automated-Retail-Checkout}.


\end{abstract}

\section{Introduction}
\label{intro}

In the rapidly evolving landscape of smart retail, the automation of processes has emerged as a pivotal endeavor, enhancing efficiency and user experience. One prominent facet of this transformation is the automation of the checkout process, wherein users seamlessly pass products through a camera-enabled environment, eliciting automated product detection, tracking, and counting. This innovative paradigm streamlines the shopping experience but also presents unique challenges, particularly in the realm of data acquisition and model training.

\begin{figure}[t] 
    \centering
    \begin{center}
        \includegraphics[width=1\linewidth]{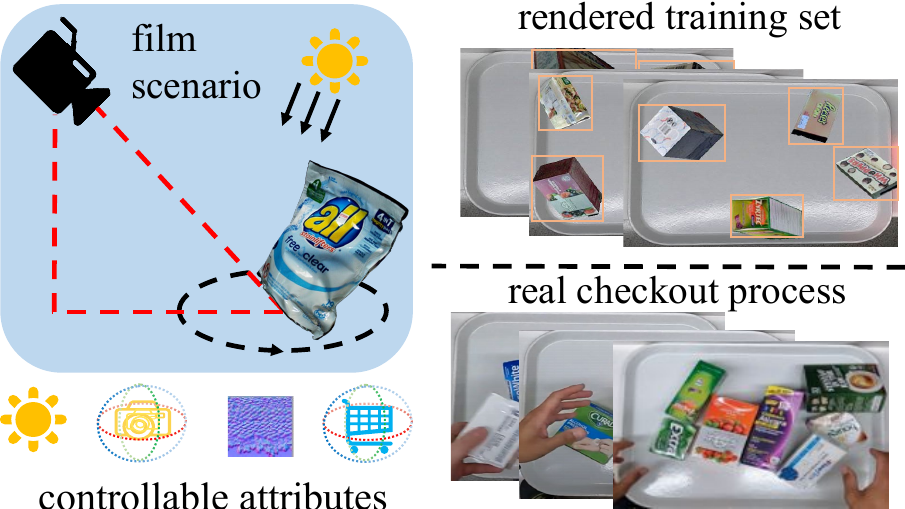}
        \caption{Problem definition. We focus on the problem of using 3D assets for training 2D detection and tracking model for AutoRetail Checkout. Given 3D assets, we aim to render a 2D training set by setting up a filming scenario. To achieve this, we propose a DtTrain framework, to improve the training set specificity on the real checkout process. 
    }
        \label{fig:prob_def}
    \end{center}
    \vspace{-2em}
\end{figure}

To achieve AutoRetail Checkout (ARC) with deep learning, the acquisition of labeled training data has become a significant bottleneck, due to its difficulty in data collection, expensive annotation, and privacy concerns~\cite{ristani2016MTMC}. For example, collecting real AutoRetail Checkout training data could be costly, as it typically involves manual product movement in front of a camera and subsequent time-consuming human labeling. In this paper, we avoid the usage of real data labeling and introduce a novel dataset that leverages product 3D models as a foundation for training data generation. Shown in  Fig.~\ref{fig:prob_def}, given 3D product models, employing a renderer allows rapid generation of rendered product images, producing a training set with thousands of images within minutes.~\cite{richter2016playing,sakaridis2018semantic,ruiz2019learning,tremblay2018training,sun2019dissecting}.

Though data rendering using graphic engines offers significant advantages for training deep learning models, it introduces a distinctive challenge, \ie, the bias difference (termed as domain gap) between rendered and real data, which hampers its scalability. To explain, as users interact with the ARC in a ``hands-on'' manner, these individual biases manifest as discernible patterns in the resulting product images. For instance, many customers tend to place the labeled side of the product facing upwards, resulting in the camera being more likely to capture this biased viewpoint. In this case, if the training data fails to encapsulate these biases, the presence of such a domain gap introduces a hurdle to the efficacy of training, as models may not generalize effectively to real-world scenarios. 

In the past, addressing domain gap and making rendered data more realistic required extensive human effort, involving complex filming scene arrangements. However, recent advancements in computer vision have brought about a revolutionary change in film scene arrangement techniques. These advancements enable training set optimization and reduce the necessity for extensive human involvement in the process ~\cite{kar2019meta,ruiz2019learning,yao2019simulating,yao2022attribute}. A prime example of these advancements is the attribute descent algorithm developed by Yao \etal. This algorithm efficiently learns film attribute distributions, significantly enhancing the realism of rendered data and improving the specificity of the training set for a target validation/testing ~\cite{yao2019simulating,yao2022attribute}. With these methods, the gap between rendered and real data is narrowed, making rendered datasets more valuable for training deep learning models.


In this paper, we present a novel pipeline for rendered training set creation by augmenting core digital twins. Firstly, our approach is motivated by the understanding that the target set bias can be effectively represented by its smaller core set. To create this core set, we carefully select the most representative samples from the target domain based on their similarity in the feature space. For representative images, we utilize the graphic engine to create their digital twins, which are virtual representations of products in the rendering environment that closely mimic real images. This is achieved through precise per-image attribute optimization with coordinate descent~\cite{wright2015coordinate}. Subsequently, we apply attribute-guided data augmentation techniques to these digital twins, thereby substantially increasing the dataset size. This data generation process culminates in the formation of the Digital Twin training set (DT set), a training dataset that specifically encapsulates the "hands-on" user bias.

\begin{figure}[t] 
    \centering
    \begin{center}
        \includegraphics[width=1\linewidth]{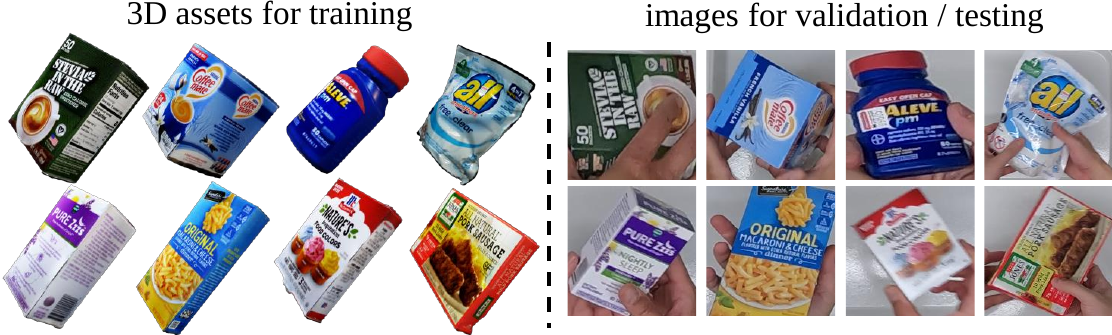}
        \caption{3D assets and real image examples in automatic retail checkout dataset. We have 3D assets for model training and 2D images for model validation and testing. 
        }
        \label{fig:auto_retail}
    \end{center}
\vspace{-1.7em}
\end{figure}

We conduct a comprehensive experiment to illustrate the superior efficacy of the DtTrain over existing training set creation methods in terms of ARC counting accuracy. Furthermore, when having joint training of the DT set and pseudo-labeled real checkout data, it results in demonstrable performance enhancements. Through this endeavor, the study offers a promising solution to the conundrum of user bias, ultimately advancing the frontier of automated checkout systems in smart retail.




\begin{table}[t]\footnotesize
\begin{center}
	\setlength{\tabcolsep}{0.8mm}{
		\begin{tabular}{c|l|c|c|c|c} 
			\Xhline{1.2pt}
			\multicolumn{2}{c|}{Datasets}	&  \#Cate. & \multicolumn{1}{c|}{\#Images} & \ Modality  & \ Attr \footnotesize \\
			\hline 
			\multirow{4}{*}{ \begin{tabular}[c]{@{}l@{}}Multi-\\modal\\Retrieval\end{tabular} }	& Dress Retrieval & 50  & 0.020M & I,T & \xmark \\
			&Product1M&458 & 1.182M & I,T  & \xmark \\
			&MEP-3M & 599 &  3.012M & I,T  & \xmark \\
                &M5Product & 6,232 & 6.313M  & I,T,V,A,Tab &\xmark  \\
			\hline
			\multirow{2}{*}{\begin{tabular}[c]{@{}l@{}}Retail\\Checkout\end{tabular}}	&RPC & 200 & 
   0.368M
   & I & \xmark \\
			& ARC (Ours)  & 116 & $\infty$  & V & \cmark \\ 
	
			\Xhline{1.2pt} 			
	\end{tabular}}
\end{center}
\caption{Comparing datasets related to retail objects. ``Attr'' denotes whether the dataset has attribute labels (\emph{e.g.,} orientation). In our ARC dataset, a category is a 3D model corresponding to a product. From each 3D model (category), we can render an unlimited number of images by varying environment and camera settings in Unity. 
Modalities are denoted as: Image (I), Text (T), Video (V), Audio (A), and Table (Tab). 
		\label{table:Datasets}}
\vspace{-1.6em}
\end{table}


\begin{figure*}[t]
\centering
\includegraphics[width=1\linewidth]{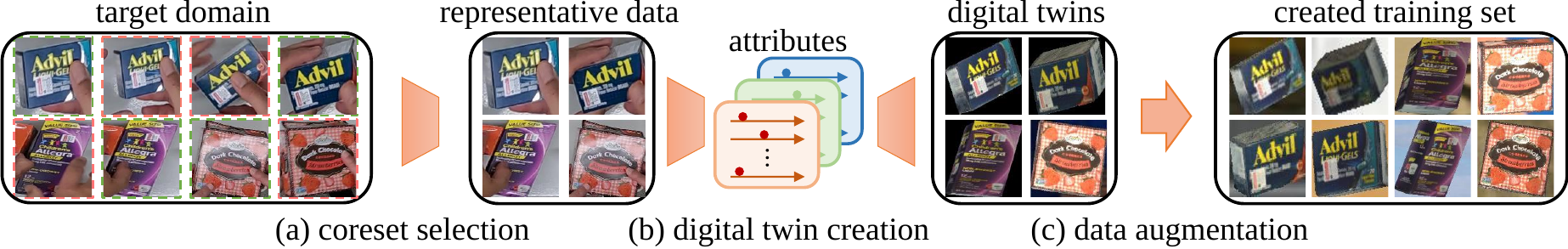}
\caption{The DtTrain framework. It is designed to construct bias-adapted rendered training data. The framework comprises three key components: (a) coreset selection, aiming at identifying the most representative samples from the target domain. In the figure, the target images with green dashed boxes are selected, and those with red dashed boxes are not selected. Following this, we (b) generate digital twins for each image within the core set, by optimizing attributes shown in Fig.~\ref{fig:prob_def}. Ultimately, the training set is curated through (c) attribute-guided data augmentation based on the rendered core set.
}
\label{fig:dataset_level_image_level}
\vspace{-1em}
\end{figure*}


\section{Method}

\subsection{Automatic Retail Checkout Dataset}
\label{sec:dataset_intro}

In this paper, the challenge lies in the absence of a labeled real-world training set for ARC. We address this by rendering different images from the given 3D retail models. The rendered data allows us to train a robust model for the 2D ARC task, even in an environment with no prior annotations for the validation/test set.

To accomplish this task, we introduce a novel dataset specifically designed for ARC. Fig.~\ref{fig:prob_def}, Fig.~\ref{fig:auto_retail} and Table~\ref{table:Datasets} illustrate examples and statistics of this dataset. We have curated 116 3D scans of real-world retail objects sourced from supermarkets, represented as 3D models. The dataset encompasses various object classes, including daily necessities, food, toys, furniture, and household items, among others. The images are captured in a setup from~\citet{yao2022attribute}. As depicted in Fig.~\ref{fig:prob_def} left, we incorporate controllable attributes like object placement, camera pose, and lighting. As highlighted in Table~\ref{table:Datasets}, compared to Dress Retrieval~\cite{corbiere2017leveraging}, Product1M~\cite{zhan2021product1m}, MEP-3M~\cite{liu2023mep}, M5Product~\cite{dong2022m5product}, and RPC~\cite{wei2019rpc}, our dataset stands out by providing 3D assets, which offer the potential to generate an extensive collection of images. Additionally, the rendered data enables us to provide accurate product labeling and further attribute labeling for real checkout scenarios. To promote collaboration and further research, we will make the 3D models and film scene (implemented by a Unity-Python interface) readily available to the community, allowing the creation of more rendered data if required. 

In a real ARC scenario, shown in the bottom right corner of Fig.~\ref{fig:prob_def}, the camera is mounted above the checkout counter and facing straight down, while a customer is enacting a checkout action by ``scanning'' objects in front of the counter in a natural manner. Several different customers participate, and each of them scan slightly differently. There is a shopping tray placed under the camera to indicate where the AI model should focus. In summary, we obtain approximately $22$ minutes of videos, and the videos are further split into target unlabeled \textit{training} and labeled \textit{test} sets such that \textit{training} and \textit{test} account for $40\%$ and $60\%$, respectively.
 
The presence of a noticeable domain gap between the rendered source and real target data is our major concern. Real-world datasets often exhibit distinct dataset biases, \eg, viewpoint bias. During a retail checkout process, customers typically view products from specific angles, resulting in uneven distribution of viewpoints. For instance, plate-like products are usually viewed from the front or rear as people handle them manually. If our rendered training set lacks a similar bias in viewpoint distribution, it creates a discrepancy between the two domains. As a consequence, the model's performance may suffer such a domain gap, leading to a drop in accuracy and effectiveness.

\subsection{Problem Definition}

Formally, we denote the \textit{target} real ARC dataset as $\mathcal{D}_T=\{({\ve x}_i,y_i)\}_{i=1}^{M}$ where $M$ indicates the number of image-label pairs in the target. It follows the distribution $p_T$, \ie, $\mathcal{D}_T\sim p_T$. Let ${\mathcal{D}_S}$ be the rendered \textit{source} set to be constructed, and  ${\mathcal{D}_S} = \{( 
\mathcal{R} (\bm{\psi_i}) ,y_i)\}_{i=1}^{N}$. Here $\bm{\psi_i}$ is an attribute vector of $K$ components controlling the 3D rendering environment, \ie, $\bm{\psi_i} = [\mu_1,...,\mu_K]\in \mathbb{R}^K$. $\mathcal{R(\cdot)}$ is the underlying rendering function that takes attribute vector $\bm{\psi_i}$ as input and produces a rendered image.  We input attribute $\bm{\psi_i}$ to the renderer that generates an image-label pair. With our renderer, we can potentially render an unlimited number of images. But here we have $N$ indicates the desired number of image-label pairs in the rendered dataset. 

With these definitions, in this paper, we aim to build ${\mathcal{D}_S}^*$ with an objective that the model $h_{\mathcal{D}_S}$ trained on $\mathcal{D}_S$ has minimized risk on $\mathcal{D}_T$, \ie, 
\begin{equation}
{\mathcal{D}_S}^* = \argmin_{\mathcal{D}_S} \mathbb{E}_{{\boldsymbol x},y \sim p_T}[\ell(h_{\mathcal{D}_S}({\ve x}), y)]. 
\label{eq:problem_define}
\end{equation}

In this paper, since we actually do not have labels in $\mathcal{D}_T$, The optimization objective we define in Eq.~\ref{eq:problem_define} is not directly tractable. Thus, we need to build ${\mathcal{D}_S}^*$ without performing real training and testing. Alternatively, we aim to get ${\mathcal{D}_S}^*$ which trains a model that can have similar performance as $\mathcal{D}_T$, which is a real target training set. Thus, formally, we transfer the objective as:
\begin{equation}
\begin{aligned}
{\mathcal{D}_S}^* &= \argmin_{\mathcal{D}_S}\big|L(h_{\mathcal{D}_T}({\ve x}), y) -
L(h_{\mathcal{D}_S}({\ve x}), y)\big|, \\
\end{aligned}
\label{eq:transformed_objective}
\end{equation}
where $L(h_{\mathcal{D}_T}({\ve x}), y)$ and $L(h_{\mathcal{D}_S}({\ve x}), y)$ are the respective risks of model $h$ on the dataset ${\mathcal{D}_T}$ and $\mathcal{D}_S$.
For explicity, we define the risk of model $h$ on an arbitrary dataset $\mathbf{S}$ as 
\begin{equation}
L(h_{{\mathbf S}}({\ve x}), y)=\frac{1}{|{\mathbf S}|} \sum_{({\ve x}_i, y_i) \in {\mathbf S}} \ell(h_{{\mathbf S}}({\ve x}_i), y_i),
\label{eq:average_loss}
\end{equation}
where $\ell(h_{{\mathbf S}}({\ve x}_i), y_i)$ is the risk on individual samples as in Eq.~\ref{eq:problem_define}. 

We further split our objective into two parts, where we aim to optimize the upper bound of the error in Eq.~\ref{eq:transformed_objective}, \ie,
\begin{align}
{\mathcal{D}_S}^* = \argmin_{\mathcal{D}_S} & \underbrace{ \big| L(h_{\mathcal{D}_T}({\ve x}), y) - L(h_{\mathcal{D}_C}({\ve x}), y) \big| }_{\text{Core Set Error}}      \nonumber
+  \\  
& \underbrace{  \big|
 L(h_{\mathcal{D}_C}({\ve x}), y) - 
L(h_{\mathcal{D}_S}({\ve x}), y)     \big| }_{\text{Digital Twin Error}}  .
\label{eq:problem_define_pruning} 
\end{align}

\subsection{Coreset Selection}

In the first part, to ensure similar performance between the model trained on $\mathcal{D}_C$ and the model trained on $\mathcal{D}_T$, we minimize the risk differences between them, \ie,
\begin{equation}
\mathcal{D}_C = \argmin_{\mathcal{D}_C \in 2^{\mathcal{D}_T}}\big|L(h_{\mathcal{D}_T}({\ve x}), y) -
L(h_{\mathcal{D}_C}({\ve x}), y)\big|,
\label{eq:coreset_selection}
\end{equation}
where $\mathcal{D}_C$ is a subset of $\mathcal{D}_T$ in the size $O$, \ie, $\mathcal{D}_C=\{({\ve x}_i,y_i)\}_{i=1}^{O}$. Thus, we reduce the problem into a core set selection problem. 

From the theory of core set~\cite{sener2018active}, if $\mathcal{D}_C$ is the $\delta$ cover of the set $\mathcal{D}_T$ and shares the same number of classes with $\mathcal{D}_T$, the risk difference between model $h_{\hat{\mathbf s}}$ and $h_{\mathcal{D}_S}$ (\ie, core set error) is bounded by
\begin{equation}
\big|L(h_{\mathcal{D}_T}({\ve x}), y) {-} L(h_{\mathcal{D}_C}({\ve x}), y)\big| {\leq} \mathcal{O}(\delta) {+} \mathcal{O}
(|\mathcal{D}_T|^{-\frac{1}{2}}).
\label{eq:coreset}
\end{equation}
$\delta$ is the radius of the cover, and $\mathcal{O}(\delta)$ is a polynomial function over $\delta$. The problem can be reduced as a K-center problem~\cite{farahani2009facility} by optimizing $\mathcal{O}(\delta)$. We apply a 2-approximation algorithm \cite{williamson2011design} to iteratively find optimal samples in $\mathcal{D}_T$ and add to $\mathcal{D}_C$. Specifically,  each optimal sample ${\mathbf z}^*$ is computed as
\begin{equation}
{\mathbf z}^* = \argmax\limits_{{\mathbf z}_i \in { \mathcal{D}_C } \setminus \mathcal{D}_T}\min_{{\mathbf z}_j \in \mathcal{D}_T} \|f({\ve x}_i)- f({\ve x}_j)\|_2, 
\label{eq:FPS}
\end{equation}
where ${\mathbf z}=({\ve x},y)$, and $f({\ve x})$ represents the feature extracted of an image ${\ve x}$. This process is named the furthest point sampling (FPS) method~\cite{eldar1997farthest}, which enables the most representative samples from a dataset to be selected iteratively until size $O$. 

\begin{figure}[t]
\centering
\includegraphics[width=0.95\linewidth]{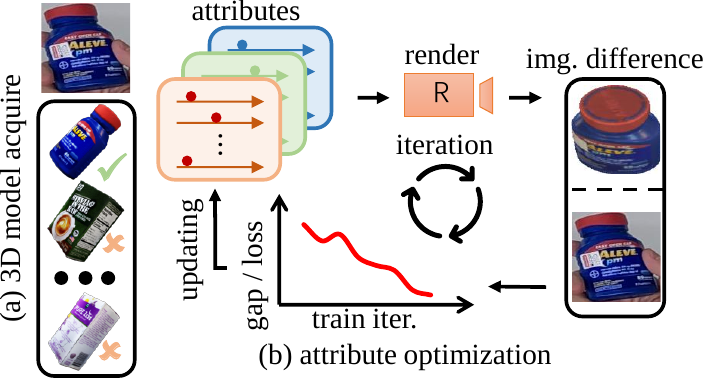}
\caption{The pipeline to obtain the digital twins. Given a 2D real image, we built its digital twin by firstly (a) acquiring the 3D assets by pseudo labeling. (b) Upon that, we render the selected 3D asset in terms of a vector of attributes, which will be iteratively optimized by coordinate descent using the image-wise difference between the rendered image and the target real image.
}
\label{fig:dataset_level_image_level}
\vspace{-1em}
\end{figure}

\subsection{Digital Twin Creation}

We then focus on the second part, we aim to get
\begin{equation}
{\mathcal{D}_S}^* = \argmin_{\mathcal{D}_S} \big|
 L(h_{\mathcal{D}_C}({\ve x}), y) - 
L(h_{\mathcal{D}_S}({\ve x}), y)     \big|.
\end{equation}
as $\mathcal{D}_C$ can be relatively small in scale (since it is a pruned set from $\mathcal{D}_T$), we consider dataset ${\mathcal{D}_S}^*$ to be identical to $\mathcal{D}_C$ by forming digital twins, by minimizing the content difference in every single image between them. Formally, we have the rendered dataset ${\mathcal{D}_S}^* = \{(\mathcal{R} (\bm{\psi_i}) ,y_i)\}_{i=1}^{O}$ and coreset $\mathcal{D}_C=\{({\ve x}_i,y_i)\}_{i=1}^{O}$. For each sample ${\ve x}_i$ in coreset $\mathcal{D}_C$, we aim to optimize $\bm{\psi_i}$, \ie,
\begin{equation}
\bm{\psi_i} ^ * = \argmin_{ \bm{\psi_i}  } \|f({\ve x}_i)- f( \mathcal{R} (\bm{\psi_i})  )\|_2, 
\label{eq:coordinate_descent}
\end{equation}
where $f(\cdot)$ denotes the feature extraction function in the feature space. In real practice, we use LPIPS~\cite{zhang2018unreasonable} to calculate image differences in feature space. 

To optimize $\bm{\psi_i}$, we are inspired by attribute descent~\cite{yao2022attribute} and use an adapted version for obtaining digital twins, \ie, coordinate descent~\cite{wright2015coordinate} for per-image optimization. Specifically, we aim to achieve the goal iteratively. Initially, at epoch $j$, we have 
\begin{equation}
\bm{\psi}_{i}^{0} = [\mu_{1}^{0}, \cdots, \mu_{K}^{0}].
\end{equation}
At epoch $j$ and iteration $k$, we iteratively optimize a single variable $\mu_{k}^{j}$, 
\begin{equation}
\begin{split}
\mu_{k}^{j} = \mathop{\arg\min}_{z \in S_{k}} \|f({\ve x}_i)- f( \mathcal{R} (\bm{\psi_i^j})  )\|_2,
\label{eq:mu}
\end{split}
\end{equation}
where 
\begin{equation}
\begin{split}
\bm{\psi_i^j} = [\mu_{1}^{j}, \cdots, \mu_{k-1}^{j}, z, \mu_{k+1}^{j-1}, \cdots, \mu_{M}^{j-1}],
\end{split}
\end{equation}
and $S_k, k=1,...,K$ defines the search space for $\mu_k$. For example, the search space for the azimuth is between $0^{\circ}$ and $330^{\circ}$ by $30^{\circ}$ degree intervals. 

In this paper, an iteration is defined as the duration for which a single attribute undergoes coordination descent optimization. 
An epoch is a duration for which all attributes undergo one attribute descent round. In this algorithm, each iteration performs a greedy search for the optimized value of an attribute while values of the other attributes are fixed. Therefore, each iteration finds the attribute value for a single attribute, and an epoch gives values for the entire attribute vector. In our experiments, the entire optimization process usually converges in 2 epochs. 

\subsection{Attribute-guided Augmentation}

From the previous step, we get a set of core digital twins ${\mathcal{D}_S}^* = \{(\mathcal{R} (\bm{\psi_i}) ,y_i)\}_{i=1}^{O}$. Though it can be used for training models directly, its dataset size $O$ can be small in size as we performed a coreset selection. To increase the dataset size to a desired number $N$, we perturb the optimized attribute values to introduce the diversity of the training set. We randomly pick an optimized attribute vector $\bm{\psi_i}^{*}$, we apply a multivariate Gaussian perturbation, denoted as $\bm{\alpha_j} \sim \mathcal{N}(\bm{\psi_i}^{*}, \Sigma)$, where $\Sigma$ is a pre-defined diagonal covariance matrix, and $i$ samples from a uniform distribution from 1 to $O$, \ie, $i \sim \mathcal{U}(1,O)$. To achieve a varied dataset resembling the digital twins, we need to strike a balance with the variance. Typically, we opt for a variance that keeps most of the values within a 15\% deviation from the mean. Given such a process, we apply augmentation multiple times to sample our final training set ${\mathcal{S}}$ (DT set) until it reaches size $N$, \ie, 
\begin{equation}
\begin{split}
{\mathcal{S}} = \{(\mathcal{R} (\bm{\alpha_j}) ,y_j)\}_{j=1}^{N}, 
\end{split}
\end{equation}
where $\bm{\alpha_j} \sim \mathcal{N}(\bm{\psi_i}^{*}, \Sigma)$, and $i \sim \mathcal{U}(1,O)$. 



\section{Experiment}

\begin{table*}[t]
\footnotesize
\centering
\setlength{\tabcolsep}{0.6mm}
\renewcommand{\arraystretch}{1.5}
\begin{tabular}{cc|cccc|cccc|cccc|cccc}
\ChangeRT{1.5pt}
\multicolumn{1}{c|}{\multirow{2}{*}{Type}} & \multirow{2}{*}{Method} & \multicolumn{4}{c|}{Bag}                                           & \multicolumn{4}{c|}{Box}                                           & \multicolumn{4}{c|}{Bottle}                                        & \multicolumn{4}{c}{All}                                          \\ \cline{3-18} 
\multicolumn{1}{c|}{}                      &                         & FID$\downarrow$             & F1$\uparrow$             & Prec.$\uparrow$          & Recall$\uparrow$         & FID$\downarrow$             & F1$\uparrow$             & Prec.$\uparrow$          & Recall$\uparrow$         & FID$\downarrow$             & F1$\uparrow$             & Prec.$\uparrow$          & Recall$\uparrow$         & FID$\downarrow$             & F1$\uparrow$             & Prec.$\uparrow$          & Recall$\uparrow$         \\ \hline
\multicolumn{2}{c|}{Random}                                          & 243.38          & 11.76          & 20.00          & 8.33           & 220.76          & 29.12          & 26.37          & 32.52          & 228.14          & 41.38          & 49.98          & 35.29          & 220.41          & 27.92          & 27.98          & 27.85          \\ \hline
\multicolumn{1}{c|}{\multirow{2}{*}{\rotatebox[origin=c]{90}{\parbox[c]{1cm}{\centering Dist. opt.}}}}                                          & LTS                     & 214.85          & 45.78          & 58.08          & 37.78          & 177.36          & 32.02          & 29.61          & 34.85          & 189.04          & 34.02          & 40.10          & 29.54          & 179.62          & 33.46          & 30.29          & 37.21          \\ \cline{2-18} 
\multicolumn{1}{c|}{}                           & Attr. Desc.             & 203.38          & \textbf{57.14} & 66.67          & \textbf{50.00} & 152.84          & 39.66          & 37.30          & 42.33          & 134.31          & 40.00          & 34.78          & 47.06          & 166.37          & 40.60          & 39.13          & 42.19          \\ \hline
\multicolumn{1}{c|}{\multirow{4}{*}{\rotatebox[origin=c]{90}{\parbox[c]{1cm}{\centering Digital Twin}}}}    & InfoGAN               & 237.50          & 50.00          & 62.50          & 41.67          & 187.48          & 30.08          & 27.55          & 33.13          & 172.10          & 25.81          & 28.57          & 23.53          & 189.42          & 30.73          & 28.90          & 32.81          \\ \cline{2-18} 
\multicolumn{1}{c|}{}                      & Soft Ras.               & 211.69          & 27.27          & 30.00          & 25.00          & 172.13          & 34.02          & 32.58          & 35.58          & 153.65          & 43.75          & 46.67          & 41.18          & 171.70          & 34.43          & 33.50          & 35.42          \\ \cline{2-18} 
\multicolumn{1}{c|}{}                      & LDM                    & 186.59          & 35.29          & 60.00          & 25.00          & 149.46          & 41.32          & 39.50          & 44.01          & 143.56          & 37.50          & 40.00          & 35.29          & 147.76          & 40.78          & 38.18          & 43.75          \\ \cline{2-18} 
\multicolumn{1}{c|}{}                      & Coor. Desc           & \textbf{177.64} & 55.56          & \textbf{83.33} & 41.67          & \textbf{128.49} & \textbf{42.32} & \textbf{40.11} & \textbf{44.79} & \textbf{122.95} & \textbf{48.48} & \textbf{50.00} & \textbf{47.06} & \textbf{128.11} & \textbf{43.43} & \textbf{42.16} & \textbf{44.79} \\ \ChangeRT{1.5pt}
\end{tabular}
\vspace{-0.5em}
\caption{Comparing different training set creation methods. We report ARC counting accuracy and FID for \textit{bag}, \textit{box}, \textit{bottle}, and \textit{all} products in different columns (all metrics are in percentage except FID). We report accuracy when training with rendered data only. }
\label{table:creation_method}
\vspace{-1em}
\end{table*}

\begin{table}[t]
\footnotesize
\centering
\setlength{\tabcolsep}{0.49mm}
\renewcommand{\arraystretch}{1.5}
\begin{tabular}{c|cc|cc|cc|cc}
\ChangeRT{1.5pt}
\multirow{2}{*}{Selection} & \multicolumn{2}{c|}{Bag}         & \multicolumn{2}{c|}{Box}         & \multicolumn{2}{c|}{Bottle}      & \multicolumn{2}{c}{All}        \\ \cline{2-9} 
                           & FID$\downarrow$             & F1$\uparrow$             & FID$\downarrow$             & F1$\uparrow$             & FID$\downarrow$             & F1$\uparrow$             & FID$\downarrow$             & F1$\uparrow$             \\ \hline
Random                     & 170.99          & 44.44          & 133.26          & 41.55          & 119.31          & 47.06          & 133.04          & 42.13          \\ \hline
FPS                    & \textbf{167.64} & \textbf{55.56} & \textbf{128.49} & \textbf{42.32} & \textbf{112.95} & \textbf{48.48} & \textbf{128.11} & \textbf{43.43} \\ \ChangeRT{1.5pt}
\end{tabular}
\vspace{-0.5em}
\caption{Comparison of different coreset selection methods. Notations and evaluation metrics are the same as in the previous table.  }
\label{table:coreset_selection}
\end{table}

\begin{table}[t]
\footnotesize
\centering
\vspace{-0.3em}
\setlength{\tabcolsep}{0.4mm}
\renewcommand{\arraystretch}{1.5}
\begin{tabular}{c|cc|cc|cc|cc}
\ChangeRT{1.5pt}
\multirow{2}{*}{Loss} & \multicolumn{2}{c|}{Bag}         & \multicolumn{2}{c|}{Box}         & \multicolumn{2}{c|}{Bottle}      & \multicolumn{2}{c}{All}        \\ \cline{2-9} 
                      & FID$\downarrow$             & F1$\uparrow$             & FID$\downarrow$             & F1$\uparrow$             & FID$\downarrow$             & F1$\uparrow$             & FID$\downarrow$             & F1$\uparrow$             \\ \hline
SSIM                 & 172.30          & 47.62          & 134.32          & 41.00          & \textbf{115.42} & \textbf{51.43} & 133.51          & 42.21          \\ \hline
StyleLoss             & 182.24          & 35.71          & \textbf{123.66} & \textbf{44.44} & 134.35          & 30.30          & 135.23          & 42.76          \\ \hline
LPIPS                & \textbf{167.64} & \textbf{55.56} & 128.49          & 43.32          & 122.95          & 48.48          & \textbf{128.11} & \textbf{43.43} \\ \ChangeRT{1.5pt}
\end{tabular}
\vspace{-0.5em}
\caption{Comparison of different loss functions. Notations and evaluation metrics are the same as in the previous table.}
\label{table:training_loss}
\vspace{-0.8em}
\end{table}

\subsection{Experiment Details}

\textbf{Task setting}. The videos are split such that 40\% of the data is to be used for target training. We are tasked to create a rendered training set by adapting to the given unlabeled target training videos, train a model on the rendered data, and report the task test accuracy on the remaining 60\% of the data, named as the target test set.

\textbf{Task model}. Once we get the optimized rendered data, we train the ARC model using the detection-tracking-counting framework as depicted from \citet{nguyen2022improving}. The pseudo-labeling model is trained from random attributes. More details are in the supplementary material. 


\textbf{Evalution metrics}. Our model evaluation entails aligning the dual outputs with the ground truth. A prediction is deemed accurate if and only if both the predicted \textit{label} and \textit{its corresponding timeframe} are correct. Specifically, we have precision, signifying the ratio of correct predictions to total predictions, and recall, reflecting the ratio of correct predictions to total ground truth. The culmination of these metrics is represented by an F1 score. Furthermore, we employ the Fr\'{e}chet Inception Distance (FID)~\cite{heusel2017gans} to assess the domain gap between the rendered set generated and the target set.

\textbf{Methods in comparison}. In this study, we conduct a comprehensive comparison between our proposed DtTrain framework and two established methods commonly employed for generating training sets through the graphic engine, namely LTS~\cite{ruiz2019learning} and attribute descent~\cite{yao2022attribute}. These methods fall under the category of attribute distribution optimization, as they necessitate the prior definition of attribute distributions and subsequent parameter optimization.

Several existing approaches for acquiring digital twins exist. Within the DtTrain framework, we compare these approaches with the coordinate descent algorithm employed in our research. Specifically, digital twin acquisition can also be achieved through the utilization of the differentiable renderer, known as the soft rasterizer~\cite{liu2019soft}. Additionally, we incorporate neural rendering techniques to procure digital twins, including InfoGAN~\cite{chen2016infogan} and latent diffusion models (LDM)~\cite{rombach2022high}. For a comprehensive understanding of the comparative methods, we provide intricate details in the supplementary material.

\subsection{Main Results}

%

\begin{figure}[t]
 \begin{minipage}[c]{0.23\textwidth}
    \includegraphics[width=\textwidth]{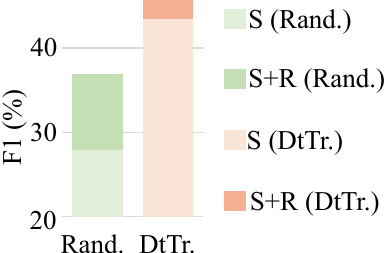}
  \end{minipage}
  \hfill
\begin{minipage}[c]{0.22\textwidth}
    \caption{Comparing training sets created by DtTrain and random attributes. 
    Two application scenarios are evaluated: training with rendered data only (``S'') and joint training on both pseudo-labeled real data and rendered data (``R+S''). 
    }
  \label{fig:randvsDtTrain}
  \end{minipage}
  \vspace{-4mm}
\end{figure}

\textbf{The superiority of DtTrain over random attributes, and existing dataset synthesis methods}. We compare the proposed DtTrain to random attributes under two settings. As shown in Fig.~\ref{fig:randvsDtTrain}, in the first setting, we only use the rendered data to train an ARC model. In the second setting, we use the rendered data combined with the pseudo-label real data to train an ARC model. Under both settings, we observe a notable superiority of DtTrain over random attributes. 

Table \ref{table:creation_method} displays evaluation results for various optimization methods, categorized by two training set creation pipelines. Notably, DtTrain outperforms existing attribute distribution optimization methods. For example, when creating digital twins with coordinate descent, the created training set surpasses the distribution optimization technique attribute descent by 2.6\% F1 score. 

Our understanding of the difference between attribute distribution optimization and our proposed pipeline aligns with the observed results. Digital twin creation involves image-to-image representations, with each product characterized by multiple augmented attribute vectors for a more intricate distribution. This highlights a key distinction that the assumption in attribute distribution optimization limits simulation potential, while our approach with multiple digital twins fosters a richer distribution. Adopting a more intuitive perspective, this image-to-image alignment eliminates interference from diverse backgrounds in digital twin creation, especially beneficial in object-centric tasks like ARC where unexpected perceptual noise can disrupt results.



\begin{figure*}[t]
\centering
\vspace{-0.5em}
\includegraphics[width=1\linewidth]{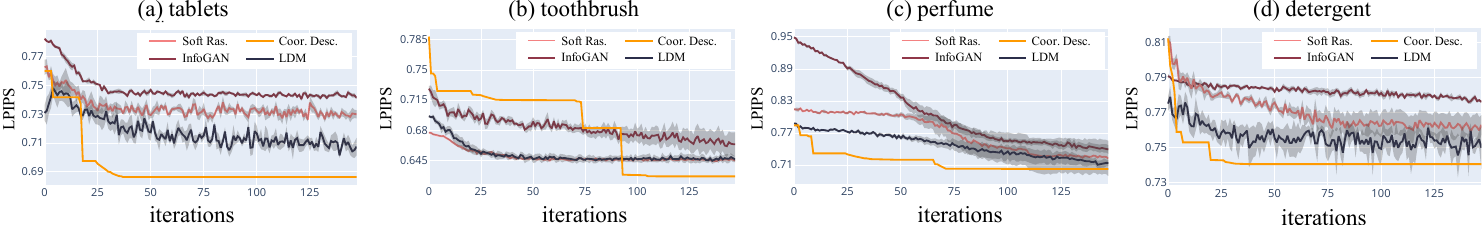}
\caption{Convergence study of digital twin creation methods. We select 4 sample products randomly from the dataset, which include 2 boxes (tablets and toothbrush), 1 bottle (perfume), and 1 bag (detergent). We plot the loss trend for each product during the optimization process using different methods: soft rasterizer, coordinate descent InfoGAN based and LDM based  neural rendering. Among these methods, coordinate descent stands out with its distinct optimization curve, characterized by a step-like descent trend. This unique behavior is attributed to coordinate descent being a search algorithm.
}
\label{fig:opt_curve}
\end{figure*}

\begin{figure*}[t]
\centering
\vspace{-0.5em}
\includegraphics[width=1\linewidth]{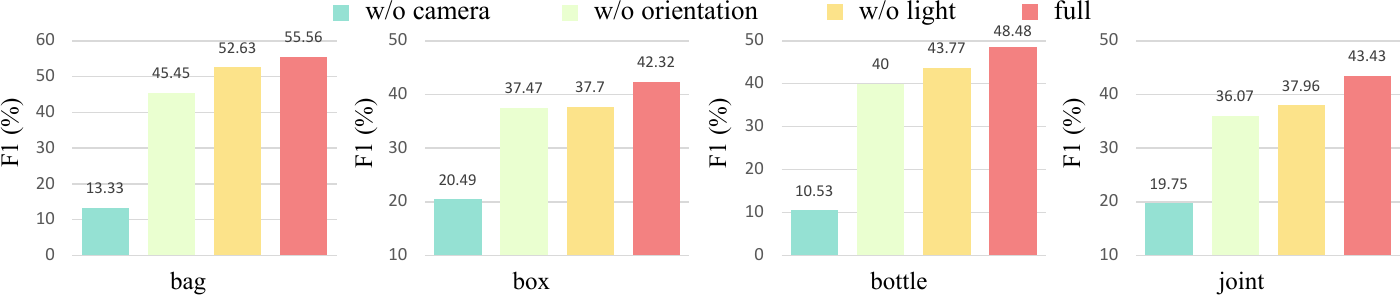}

\caption{Ablation study on attributes. We divide the attribute vector into camera (height and distance), orientation (azimuth and in-plane rotation), and light (intensity). By separately analyzing the role of each attribute group in optimization and comparing it to the full optimization, we can assess their individual impact. The compromise of task accuracy (F1) serves as an indicator of the relative importance of the isolated attribute group.
}
\label{fig:ablation_study}
\vspace{-1em}
\end{figure*}

\begin{figure}[t] 
    \centering
    \begin{center}
        \includegraphics[width=1\linewidth]{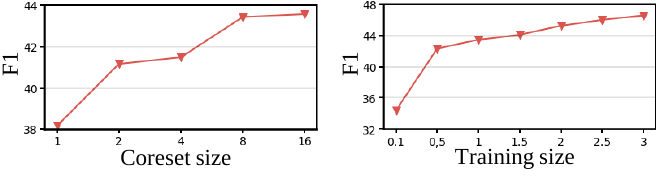}
        \caption{The parameter study of the coreset size (\textbf{Left}) and training size (\textbf{Right}). It exhibits the task accuracy trend with the increment of the indicated parameter.
        }
        \label{fig:parameter_study}
    \end{center}
\vspace{-1em}
\end{figure}

\begin{figure*}[t]
\centering
\includegraphics[width=1\linewidth]{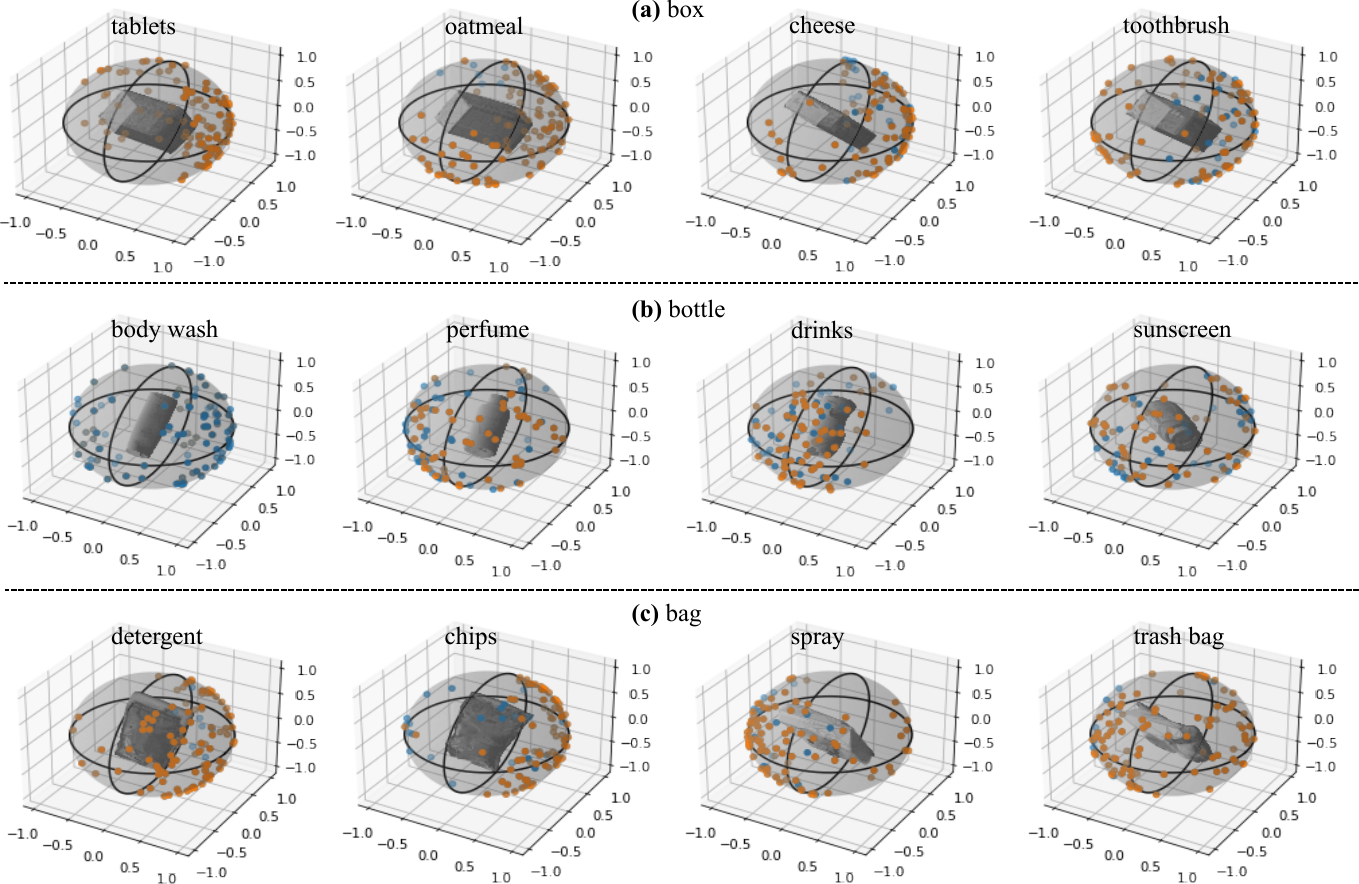}
\vspace{-0.3em}
\caption{Viewpoint distribution visualization for box, bottle, and bag retail products. Viewpoint distributions are learned by DtTrain. We select 4 products per class, where blue samples imply the in-plane rotation under $30^\circ$ and orange for above. Each sample, represented by a point surrounding the object, corresponds to learned attributes by DtTrain.  Remarkably, the distribution exhibits a non-uniform pattern, resulting in noticeable biases within each class. Our proposed method aims to model the intricate attribute patterns to accurately simulate the biases such that the domain gap is minimized.
}
\label{fig:viewpoint}
\vspace{-1em}
\end{figure*}

\textbf{Enhanced accuracy through joint training with pseudo-labeled target data}. As depicted in Fig.~\ref{fig:randvsDtTrain}, our findings underscore the substantial advancements achieved via joint training in comparison to employing solely rendered data for training purposes. Notably, upon amalgamating the DT set with pseudo-labeled authentic data, a notable 2.21\% enhancement in F1 score is observed, signifying the pronounced efficacy of this joint approach over the exclusive use of the DT set.


\textbf{The superiority of coordinate descent over existing digital twin creation methods}. In Fig. \ref{fig:opt_curve}, the risk curve spans 150 iterations, illustrating convergence patterns of optimization methods, including soft rasterizer, InfoGAN, LDM-based, and coordinate descent. Notably, coordinate descent exhibits a stable convergence, which is a distinctive step-like descent, in contrast to gradient-based behaviors which are usually not stable. This distinction stems from coordinate descent's search strategy. These observations align with our previous assessment in Table \ref{table:creation_method}. Coordinate descent consistently outperforms others, in terms of both domain gap and the final ARC accuracy.


\textbf{The superiority of FPS over random sampling}. Coreset selection in digital twin creation is crucial, capturing representative images from a large target image pool, thereby reducing the number needed for creating digital twins. To validate the effectiveness of our coreset selection method FPS, we compare it to random selection. In the latter, an equivalent number of images are randomly chosen from the target set. Results in Table \ref{table:coreset_selection} highlight FPS's clear superiority. It exhibits higher task accuracy and domain gap, outperforming random selection by 1.3\% and 4.93, respectively. 

\textbf{The superiority of LPIPS}. 
In our experiment, we utilize LPIPS as our loss function for digital twin creation. To gain deeper insights, we explore alternative losses: SSIM~\cite{wang2004image}, and StyleLoss~\cite{johnson2016perceptual}. Results in Table \ref{table:training_loss} show LPIPS as the superior choice for domain dissimilarity and task accuracy. 


\textbf{Impact of different attributes}. In our film scene, we group the attribute vector into three categories, the camera (distance, height), orientation (in-plane rotation, azimuth), and light (intensity). The ablation of each attribute can reveal their impact on task accuracy. In Fig. \ref{fig:ablation_study}, 
notably, we observe camera location (distance, height) holds a dominant role in domain dissimilarity. Distant objects exist in object-centric tasks, leading to lower resolution and quality. Lighting and orientation exert similar influences, with a slight orientation advantage due to viewpoint distribution bias.

\textbf{Parameter study}. By default, we select 8 target images per product. Thus the size of ${\mathcal{D}_S}^*$, $O = 8 \times 116$. Increasing corset size $O$ enhances target distribution representation but escalates attribute optimization time. Our experiment, depicted in the left of Fig. \ref{fig:parameter_study}, reveals an evident trend, \ie, larger coreset sizes enhance task accuracy. However, larger coreset sizes also increase the time needed for building a training set.  This highlights a trade-off, emphasizing the need to balance accuracy near saturation while maintaining operational efficiency when selecting an optimal coreset size. We observe that beyond a coreset size of 8, accuracy improvement plateaus. Thus by default, 8 target images per product are selected.


In the experiment, we have defaulted the size of a training set $N$ equals 22,000. We also test different training set sizes relative to the default size. Results, shown on the right of Fig. \ref{fig:parameter_study}, clearly indicate increasing training size enhances task accuracy. However, accuracy improvement plateaus when it reaches the default training set size.

\textbf{Numerically understanding real ARC bias}. The distribution of viewpoints is illustrated in Fig. \ref{fig:viewpoint}, providing valuable insights into inherent biases. By establishing a correlation between viewpoint distribution bias and the underlying shape characteristics, we can intuitively elucidate the rationale behind these patterns. For instance, the unimodal distribution observed in box-like tablet products indicates a customer preference for holding the product from a specific angle. 
In comparison, the nearly uniform distribution of body wash can be attributed to the cylindrical symmetry of its bottle-like shape. A bimodal distribution emerges for bag-like objects such as chips, showing customers' viewpoints concentrated at the front and back of the bag.


\section{Conclusion}

In conclusion, the automation of the checkout process in smart retail environments has garnered significant attention. However, the scarcity of annotated training data has posed a challenge. To overcome this limitation, we introduced a novel approach utilizing product 3D models for data generation through graphic engine rendering. This approach, termed as  DtTrain, can automatically edit the rendered image content in a graphic engine to generate training data with a good resemblance to the real ARC scenario. In addition, using viewpoint as an example, we show that models enable understanding of the dataset (user) bias computing the attribute distribution of given product categories. This article demonstrates the benefit of training data optimization, and establishes a promising pathway for advancing automated checkout systems in smart retail through robust and representative training data.

\section{Acknowledgement}

This work is partially done when Yue has an internship at
NVIDIA, with the support of NVIDIA computing resources. This work was also supported in part by the ARC Discovery Project (DP210102801), Oracle Cloud credits, and related resources provided by Oracle for Research.


\section{A \quad Related Works}

\textbf{AutoRetail checkout} has been advanced through multimodality, initially relying on barcodes~\cite{sriram1996applications}, a technology that remains widely used and popular today. Recent developments in deep learning have sparked a shift in ARC research, with a growing emphasis on computer vision approaches. Notably, researchers have explored the utilization of VGG-16 and Inception V3 layers as feature descriptors for image classification of various products~\cite{geng2018fine, chong2016deep}. 
Additionally, deep learning pipelines based on state-of-the-art object detectors have been proposed for product recognition~\cite{tonioni2018deep}. In contrast, our focus diverges from previous work that primarily emphasized model design and tuning. Instead, we concentrate on rendered image creation and optimization, aligning with the downstream task and target domain.

\textbf{Training with rendered data.} Data rendering via graphic engine has emerged as a cost-effective alternative to real labeled data, which can often be expensive to obtain. Various studies have explored the integration of rendered data alongside real data in the training set to improve model accuracy~\cite{yao2019simulating, zheng2017unlabeled}. Additionally, some researchers have delved into training models exclusively on rendered data~\cite{kar2019meta}. In the context of this paper, we concentrate on leveraging rendered data exclusively to develop automatic retail systems.

\textbf{Existing training set optimization methods.} 
Leveraging rendered data offers the advantages of increased label availability and enhanced flexibility in the training process. However, it also introduces challenges, most notably the domain gap that exists between the source and target domains, thereby reducing task accuracy. Many existing studies employ reinforcement learning (RL) to improve task accuracy by optimizing rendering attributes~\cite{kar2019meta, devaranjan2020meta, ruiz2019learning, xue2021learning}. For instance, Kar \etal use policy gradients to optimize scene layout. 
In comparison, Yao \etal formulate attribute optimization as a search problem due to challenges in obtaining attribute gradients, proposing a pruned greedy search called attribute descent~\cite{yao2019simulating}. However, these existing methods still require the manual definition of attribute distributions, which involves significant human effort. In this paper, we introduce the DtTrain framework, which minimizes the need for human-designed attribute distributions while achieving higher task accuracy. With DtTrain, the burden of distribution design is reduced, providing a more efficient and effective approach to attribute optimization.


\begin{figure}[t]
\centering
\includegraphics[width=0.75\linewidth]{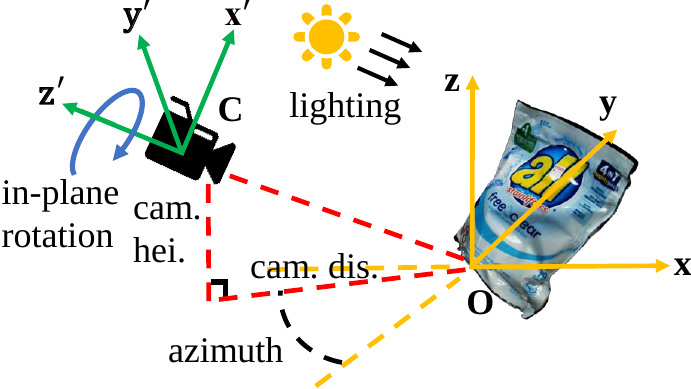}
\caption{Controllable attributes defined in our film scene. Inherited from the film scene proposed by~\citet{yao2022attribute}, we have the attributes that include in-plane rotation, azimuth, camera distance, camera height and lighting. }
\label{fig:attribute_overview}
\vspace{-1em}
\end{figure}

\section{B \quad Film Scene}
We consider that the content disparity in images arises from various prime factors. In this context, a rendered image is created through a renderer, which is conditioned on attributes including azimuth, in-plane rotation, lighting intensity, lighting angle, camera distance, and height. These attributes are defined in the work by \citet{yao2022attribute} and are explained in Fig. \ref{fig:attribute_overview}. This supplementary figure complements the information presented in Fig. \ref{fig:prob_def}, providing a more detailed understanding of the physical significance of these attributes.

We first constrain the rotation attributes, namely in-plane rotation, azimuth, and light direction, to fall within the range of $0^\circ$ to $360^\circ$. Subsequently, the light intensity is deliberately designed to vary between 0 and 100. Here, a value of 0 signifies complete darkness, while a value of 100 corresponds to full illumination. Regarding camera height and distance, we refrain from explicitly specifying a range limit, varying between 0 and 100. 
Since we use coordinate descent, a search-based technique for obtaining digital twins. To accommodate this, we define search spaces $\bm S = [S_1, \cdots, S_K]$ for each attribute in the attribute list $\bm \psi = [\psi_1, \cdots, \psi_K]$. For example, we define the search space of rotation attributes, which lies in the range of $0^\circ$ to $360^\circ$ that has $30^\circ$ degrees intervals. For the search space of camera height and camera distance, we have a range of $0$ to $100$ that has $10$ intervals. In the context of coordinate descent, a balance between interval size and search space emerges, \ie, small intervals correspond to expansive search spaces, while larger intervals narrow down exploration. To ensure experimental fairness, we maintain a consistency akin to \citet{yao2022attribute}.


\section{C \quad Existing Optimization Methods}
We compare our methods with various existing optimization strategies from previous research in the experiment. As mentioned in our experiment part, we compare DtTrain with existing training set optimization methods, and we compare coordinate descent with existing digital twin creation methods. 

\subsection{Existing Training Set Optimization Methods}

We utilize two well-established techniques frequently employed for creating training datasets using the graphical engine. These techniques are known as LTS \cite{ruiz2019learning} and attribute descent \cite{yao2022attribute}. These approaches belong to the realm of attribute distribution optimization, as they require the initial specification of attribute distributions followed by subsequent optimization of parameters. 

\textbf{Learning to Simulate (LTS)}~\cite{ruiz2019learning} is a typical distribution-based reinforcement learning approach, where the controllable attributes are optimized by maximizing the reward accumulated in the downstream task. Despite the simple design, the end-to-end learning architecture implies it is difficult to combine with complicated downstream tasks. In our experiment, we regard it as an ad-hoc benchmark in conventional distribution optimization approaches.

\textbf{Attribute descent}~\cite{yao2022attribute} is a gradient-free search strategy to get an optimized training set. During the search stage, it can 
significantly reduce the search space of the distribution parameters. Furthermore, attribute descent is guaranteed to find the sub-optimal solutions by iterating a single element while keeping others frozen.

\begin{figure}[t]
\centering
\includegraphics[width=0.87\linewidth]{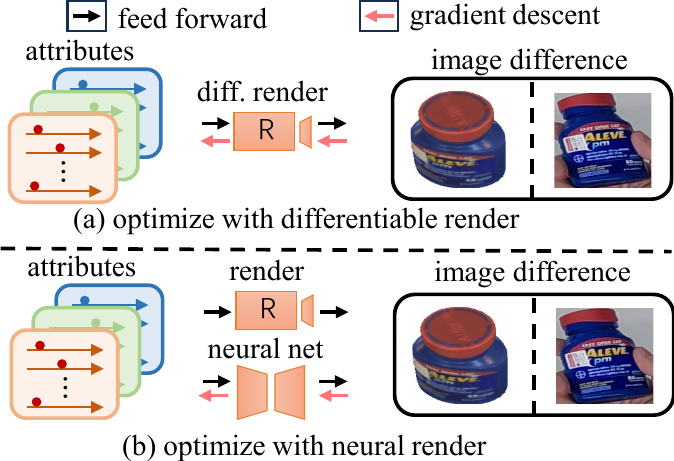}
\caption{ Existing two methods for creating digital twins, differentiable rendering and neural rendering. For the differentiable renderer, the gradient can directly backpropagate through the renderer to optimize the attributes. For the neural renderer, we create a conditional neural network to emulate differentiable renderers.
}
\label{fig:pipeline3}
\vspace{-1em}
\end{figure}
\subsection{Existing Digital Twin Creation Methods}

For both differentiable rendering and neural rendering, we use the same loss as coordinate descent, LPIPS. 

\textbf{Differentiable rendering} refers to a renderer that preserves gradients in the rendering function, shown in Fig. \ref{fig:pipeline3} top. In our experiment, we use a differentiable renderer called soft rasterizer~\cite{liu2019soft}, which is implemented in Pytorch3D~\cite{ravi2020pytorch3d}. In the soft rasterizer, the gradient can directly backpropagate through the renderer, thereby enabling the editing of attributes to create digital twins. 


\textbf{Neural rendering} can be seen as a similar form of differentiable renderer as it also enables gradient-based optimization.
This approach performs backpropagation through the neural network, which serves as an imitator of a non-differentiable renderer. This approach is originally proposed by \cite{shi2019face} to create digital twins for face images. Motivated by their methods, we have an adapted version to create digital twins for product images. Specifically, as shown in Fig. \ref{fig:pipeline3} bottom. The whole process involves two stages. In the first stage, we train the conditional generative neural network using the attributes and their corresponding rendered images. In the second stage, we freeze the network while backpropagating gradients through the neural renderer, thereby enabling the editing of attributes to create digital twins.  

In the experiment, we utilize two typical conditional generative networks: InfoGAN~\cite{chen2016infogan} and latent diffusion model (LDM)~\cite{rombach2022high}. By modifying specific tailored latent variables, the InfoGAN can generate product images with desired attributes. Likewise, LDM is widely popular due to its strong representation capabilities and high-quality outputs. In our experiment, we adopt the pre-trained stable diffusion model from \cite{rombach2022high} and fine-tune it on our created retail training set. As the original architecture only supports text prompts, we make adjustments by removing the CLIP encoder~\cite{radford2021learning}, and replacing it with a lightweight, zero-initialized attribute encoder. For LDM training, we train the entire U-Net~\cite{ronneberger2015u} along with the attribute encoder to adapt the model to our retail products.

As shown in our experimental results, we demonstrate that coordinate descent consistently outperforms both differentiable rendering and neural rendering. While both differentiable rendering and neural rendering involve the iterative update of attributes through gradient descent. It is noteworthy that, in contrast to coordinate descent, these gradient descent-based approaches are susceptible to becoming entrapped within local minima. To illustrate, consider a nearly symmetrical retail product, such as a box-shaped item, where viewpoint local minima exist both on the product's front and rear sides. Despite the global minimum residing exclusively on the product's front side, gradient descent-based methods can easily become ``trapped'' in the local minimum of the rear side. In comparison, coordinate descent exhibits the ability to escape such entrapment due to its inherent search-based nature.

\section{D \quad Task Model}
We provide a comprehensive overview of the task model employed in our experiment for evaluating rendered retail data. The ARC pipeline, as encapsulated by \citet{naphade20226th}, unfolds as three distinct stages: detection, tracking, and counting (DTC).

Initiating with the detection stage, our objective revolves around the identification and classification of target products within video content. To this end, we harness the power of YOLOv5~\cite{glenn_jocher_2022_7347926}, which is pertained on COCO~\cite{lin2014microsoft} and further fine-tuned using our rendered data. The outcome manifests as estimated bounding boxes, and the classification process culminates in an ensemble of three models: Res2Net~\cite{gao2019res2net}, Swin-Transformer~\cite{liu2021swin}, and RepVGG~\cite{ding2021repvgg}. 


Transitioning seamlessly to the tracking stage, our focal point lies in establishing continuity in tracking identical products across diverse frames. This mitigates the risk of count duplication. We use the tracking algorithm, \ie, ByteTrack~\cite{zhang2022bytetrack}, which aptly achieves track matching and effectively circumvents the challenge posed by object occlusion.
Finally, in the counting stage, the goal is to predict each product uniquely, avoiding any redundancy. 
To this end, trajectory counting is employed, with each trajectory being assigned a time prediction rooted in the highest level of confidence.

\bibliography{aaai24}

\end{document}